\documentclass[letterpaper, 10 pt, conference]{ieeeconf}  

\IEEEoverridecommandlockouts                              

\usepackage{graphicx}%
\usepackage{multirow}%
\usepackage{amsmath,amssymb,amsfonts}%
\usepackage{mathrsfs}%
\usepackage{xcolor}%
\usepackage{textcomp}%
\usepackage{manyfoot}%
\usepackage{booktabs}%
\usepackage{algorithm}%
\usepackage{algorithmicx}%
\usepackage{algpseudocode}%
\usepackage{listings}%
\usepackage{caption}
\usepackage{subcaption}
\usepackage{svg}
\usepackage{stfloats}
\usepackage{float}
\usepackage{placeins}
\usepackage{cite}
\usepackage{hyperref}

\usepackage{url}
\title{\LARGE \bf
Voxel Map to Occupancy Map Conversion Using Free Space Projection for Efficient Map Representation for Aerial and Ground Robots
}

\author{Scott Fredriksson$^*$, Akshit Saradagi and George Nikolakopoulos%
\thanks{$^*$ Scott Fredriksson is the corresponding author (email : scofre@ltu.se).
The authors are with the Robotics and AI group, in the Department of Computer Science, Electrical and Space Engineering at Luleå University of Technology, Sweden.}%
}

\begin{document}

\maketitle
\thispagestyle{empty}
\pagestyle{empty}

\begin{abstract}
This article introduces a novel method for converting 3D voxel maps, commonly utilized by robots for localization and navigation, into 2D occupancy maps for both unmanned aerial vehicles (UAVs) and unmanned ground vehicles (UGVs). The generated 2D maps can be used for more efficient global navigation for both UAVs and UGVs, in enabling algorithms developed for 2D maps to be useful in 3D applications, and allowing for faster transfer of maps between multiple agents in bandwidth-limited scenarios.
The proposed method uses the free space representation in the UFOMap mapping solution to generate 2D occupancy maps. 
During the 3D to 2D map conversion, the method conducts safety checks and eliminates free spaces in the map with dimensions (in the height axis) lower than the robot's safety margins. This ensures that an aerial or ground robot can navigate safely, relying primarily on the 2D map generated by the method.
Additionally, the method extracts the height of navigable free space and a local estimate of the slope of the floor from the 3D voxel map. The height data is utilized in converting paths generated using the 2D map into paths in 3D space for both UAVs and UGVs. The slope data identifies areas too steep for a ground robot to traverse, marking them as occupied, thus enabling a more accurate representation of the terrain for ground robots. The effectiveness of the proposed method in enabling computationally efficient navigation for both aerial and ground robots is validated in two different environments, over both static maps and in online implementation in an exploration mission. 
The methods proposed within this article have been implemented in the popular robotics framework ROS and are open-sourced. The code is available at: \href{https://github.com/LTU-RAI/Map-Conversion-3D-Voxel-Map-to-2D-Occupancy-Map}{https://github.com/LTU-RAI/Map-Conversion-3D-Voxel-Map-to-2D-Occupancy-Map}.
\end{abstract}

\section{INTRODUCTION}\label{sec:introduction}
A map is a fundamental component of robotics, serving key roles in localization, trajectory planning, and obstacle avoidance. The most common form of map representation is the metric grid-based map, in either 2D or 3D formats. Both 3D and 2D grid-based maps have their respective advantages: 3D maps are more accurate representations of the world and offer greater detail and precision, while 2D maps offer the advantage in terms of performance and minimality, particularly with respect to computational efficiency and memory usage. Therefore, utilizing both representations can be beneficial \cite{Megalingam2023} for path planning in autonomous robots. A common approach involves creating and managing two separate 2D and 3D mapping solutions, 
by either converting 3D scans used in 3D mapping to 2D scans for creating the 2D map \cite{Wulf2004,Sandfuchs2021,Mora2023} or using multiple sensors \cite{Nam2017}. However, this leads to the challenge of managing two different mapping frameworks on a single robot, potentially resulting in mismatched maps. This paper aims to effectively integrate the distinct mapping advantages of 2D and 3D maps by proposing a method for converting 3D voxel maps into 2D maps to enable efficient path planning for both unmanned aerial vehicles (UAVs) and unmanned ground vehicles (UGVs).

In the robotics literature, there are very few solutions for effectively converting 3D voxel maps into 2D maps. The most prevalent technique involves projecting a part of the voxel map at a fixed height downward onto a 2D plane, a process implemented in the widely-used OctoMap library within ROS (Robot Operating System) \cite{Hornung2013}.
Another example is the hybrid height voxel mapper, HMAPs~\cite{Yang2018}, which allows for the conversion from 3D to 2.5D and 2D maps using the same downward projection technique.
Such approaches have several limitations. Specifically, the projected part of the map do not include any voxels from the ground or ceiling, as these would also be projected onto the 2D map. Similarly, any obstacle in the map that is not within the selected range will be omitted from the projection. Generally, this is not a problem in smaller, controlled environments, but in larger environments with varying height of navigable space, with multiple height levels or in cases where there is drift in the robot's positioning along the vertical axis, this method becomes less effective. 

While there are not many solutions for converting voxel-based 3D maps into 2D occupancy maps, several methods suitable for point clouds do exist. In the work \cite{Li2024}, down projection is utilized to convert a 3D point cloud into a 2D map, after filtering out the portion of the point cloud that represents the floor.
In the past, multiple studies have also explored the conversion of point clouds generated by Kinect cameras ~\cite{Kamarudin2013,Garrote2017,Brahmanage2019} into 2D maps. However, such methods are unsuitable for 
handling point clouds containing ceilings or overhangs. 
The method in \cite{Yusefi2020} generates a 2D map using keyframes from ORB-slam \cite{MurArtal2015}, while the solution in ~\cite{Yang2018,Gim2021} uses 2.5D mapping, focusing on mapping only surfaces that a ground robot can traverse.
%
%
\section{Contributions} \label{sec:contribution}
The article's primary contribution is a novel method for converting a 3D voxel map into a 2D occupancy map. 
Instead of projecting a fixed section of the voxel map, as in existing methods, the proposed method projects free space, leading to a more robust and flexible approach in environments with varying elevations. The method also extracts the height values of the ceiling and floor of the environment and a local estimate of the slope of the floor. 
The proposed method generates two 2D maps: i) an obstacle-free map that is useful for computation-friendly UAV path planning, by taking into account the height of the floor and ceiling from the height map, and ii) a map for UGVs that incorporates walls and obstacles using slope estimation, which again enables more computation-friendly path planning for UGVs in comparison with 3D Voxel maps. 
The second contribution is the proposal of a method to convert 2D paths generated using 2D occupancy maps into 3D paths for UAVs and UGVs using the height map. This allows an aerial or ground robot to navigate safely, relying primarily on the 2D map generated by the method. 
These methods have been implemented in ROS and open sourced\footnote{Source code is available at \href{https://github.com/LTU-RAI/Map-Conversion-3D-Voxel-Map-to-2D-Occupancy-Map}{https://github.com/LTU-RAI/Map-Conversion-3D-Voxel-Map-to-2D-Occupancy-Map}}. The proposed method is implemented using the UFOMap Mapping framework \cite{Duberg2020}. 
However, the method is compatible with any voxel-based mapping solution that represents space using occupied, free, and unknown voxel cells. 
%
%

\begin{figure*}[t]
    \vspace{10pt}
    \centering
    \includegraphics[width=\linewidth]{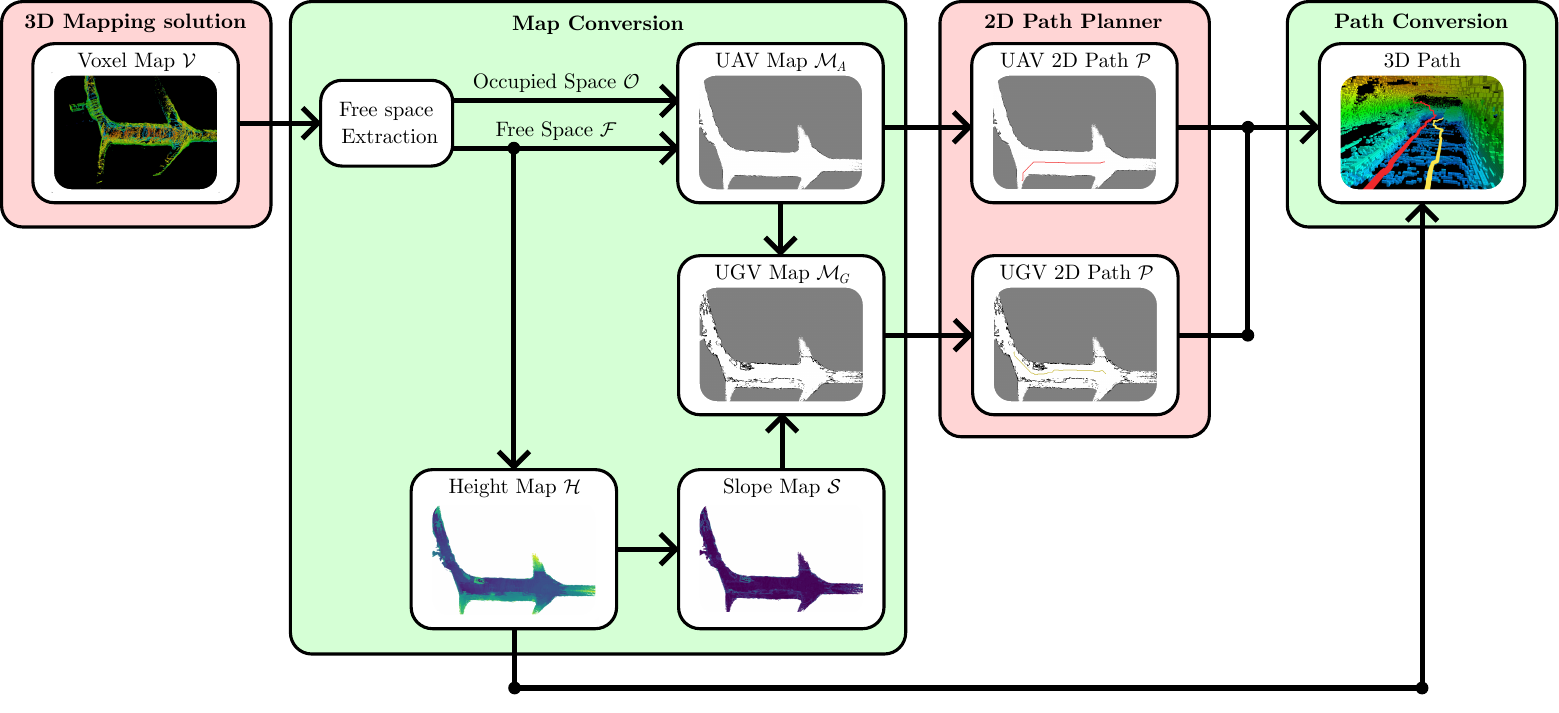}
    \caption{An overview of the proposed methodology for 3D Voxel map to 2D map conversion, along with path planning for the UAVs and UGVs, and 2D to 3D path conversion.}
    \label{fig:overview}
\end{figure*}
The rest of the  article is structured as follows. 
Section \ref{sec:method} describes the proposed methodology for 3D to 2D map conversion. Section \ref{sec:path} demonstrates the implementation of the methodology for converting 2D paths into 3D paths using the height map generated by the map conversion process. Section \ref{sec:useCases} discusses the utility of the 3D to 2D map conversions in robotics applications. Section \ref{sec:validation} presents the validation of the overall methodology on maps from two different environments. Section \ref{sec:discussion} contains a discussion of the results. Finally, Section \ref{sec:conclusion} presents conclusions, with a discussion and summary of the findings.
%
%
\section{3D to 2D Map Conversion}\label{sec:method}
An overview of the proposed methodology for 3D Voxel map to 2D map conversion is presented in Figure \ref{fig:overview}. Using the 3D voxel map $\mathcal{V}$, four 2D grid-based maps are generated. Firstly, a height map $\mathcal{H}$ is created, where every cell of a 2D grid map contains height data of the navigable free space (ceiling and floor) relative to the origin of the Voxel map. Secondly, a slope map $\mathcal{S}$ is generated from the height map (floor height) by estimating the slope of the floor plane in the immediate neighborhood of every cell. 
Finally, two occupancy maps $\mathcal{M}_A$ and $\mathcal{M}_G$ are produced, one for the UAVs and the other for UGVs respectively, assigning values: $-1$ for unknown cells, and a value in the range $[0,1]$ for the occupancy probability, with $0$ being free, and $1$ being occupied. All 2D maps are generated over a 2D grid $\mathcal{G}_{2D}$ with the same $x$ and $y$ size/extent and resolution $\mathcal{V}_{res}$ as the Voxel map being converted. 
Details about the individual steps in the methodology are presented next.

\subsection{Extraction of Free Space and Height Map} \label{sec:FreeSpace}
Most robots have their LiDAR mounted vertically, resulting in an incomplete view of the floor. This configuration leads to numerous gaps in the mapped floor area, as can be seen in Figures \ref{fig:caveV1} and \ref{fig:caveV2} and Figures \ref{fig:outdoorV1} and \ref{fig:outdoorV2}. To address this issue, instead of using occupied voxels, the free space in the map is utilized to identify the floor in $\mathcal{V}$.

\subsubsection{Free and occupied space} To identify the floor, the map $\mathcal{V}$ is first represented as a union of free $(\mathcal{F})$ and occupied $(\mathcal{O})$ ranges on every cell $\{x_m,y_n\}$ in the 2D grid $\mathcal{G}_{2D}$ of size $M \times N$. The free space in the 3D map is represented by the set $\mathcal{F} = \{F_{x_m,y_n}\}_{M \times N}$, with the element $F_{x_m,y_n}=\{f_1,f_2,\dots,f_{k_{mn}}\}$, where each $f_i \in F_{x_m,y_n}$ is a range of free space along the height axis, at 2D grid location $\{x_m,y_n\}$ and $k_{mn}$ is the number of such free space ranges. 
Each range of free space $f_{i}$ contains two values, 
$f_{i}=\{\Hat{f}_{i}, \Check{f}_{i}\}$, where $\Hat{f}_i$ and  $\Check{f}_i$ are the higher and lower bounds of the height range respectively. Similarly, the occupied space is represented as 
$\mathcal{O}=\{O_{x_m,y_n}\}_{M\times N}$
where $O_{x_m,y_n}=\{o_1, o_2,\dots,o_{l_{ij}}\}$ is the set of occupied ranges $o_{i}=\{\Hat{o}_{i}, \Check{o}_{i}\}$. The ranges in the set $F_{x_m,y_n}$ and in the set $O_{x_m,y_n}$ cannot overlap or be connected with another range in the same set. 
Since a 2D path planning algorithm utilizing the 2D maps generated in this work cannot perform collision checks on the height axis (the $Z$ axis), the proposed method disregards any free range $f_i \in \mathcal{F}$ with a height less than the safety margin $R_{maxZ}$, which is dependent on the dimensions of the robot. 
In other words, an $f_i$ is removed if $|f_{i}|= (\Hat{f}_{i}-\Check{f}_{i}) < R_{maxZ}$, where $|f_{i}|$ denotes the height of the range.
\subsubsection{Height map} 
Given the free space ranges in $\mathcal{F}$, the height map $\mathcal{H}=\{h_{x_m,y_n}\}_{M\times N}$ 
is built as a collection of height ranges $h_{x_m,y_n}$, one for each cell $\{x_m, y_n\}$. An element of the height map $h_{x_m,y_n}=\{\Hat{h}_{x_m,y_n}, \Check{h}_{x_m,y_n}\}$, where the bottom of the height range $\Check{h}_{x_m,y_n}$ is the height of the floor and $\Hat{h}_{x_m,y_n}$ is the height of the ceiling with respect to the origin of the voxel map. 
The floor $\check{h}_{x_m,y_n}$ of a cell $h_{x_m,y_n}$ is the bottom of $f_1 \in  F_{x_m,y_n}$, and the ceiling $\hat{h}_{x_m,y_n}$ is  the top of $f_{k_{mn}} \in F_{x_m,y_n}$. The above formulation assumes only one level of navigable space in the map in the height axis, as in a building with just one floor or a subterranean environment with no overlapping tunnels. 
%
%
\subsection{Derivation of the Slope Map through Least Squares} \label{sec:slope}
The slope map, denoted as $\mathcal{S}=\{s_{x_m,y_n}\}_{M \times N}$ is a 2D grid where each cell $s_{x_m,y_n}$ contains an estimate of the slope of the floor in a neighborhood of the cell $s_{x_m,y_n}$, which is the slope of the plane fitted using the floor height $\Check{h}_{x_{i}, y_{j}}$ and position $\{x_{i}, y_{j}\}$ of the cells neighboring the position $\{x_m, y_n\}$ in the grid. 
The cells $h_{x_i,y_j}$ that satisfy the condition $\max(|x_m-x_i|,|y_n-y_j|) \leq S_A$ are considered when calculating the slope of the cell $s_{x_m,y_n}$. 
$S_A \in \mathbb{N}_{\geq 1}$ is a tunable parameter for the size of the neighborhood considered when calculating the slope.
To fit a plane at the cell $s_{x_m,y_n}$, the linear equation of the form
\begin{equation}
    ax+by+c=z
    \label{eq:plane}
\end{equation}
is considered, where the parameters $a$, $b$ and $c$ are real numbers. The interest here is in finding the slope of a plane that is a best fit for the height and position of the cells neighboring the cell $s_{x_m,y_n}$, that is 
\begin{equation}
    \underbrace{\begin{bmatrix}
        x_{i_1} & y_{j_1} & 1 \\
        x_{i_2} & y_{j_2} & 1 \\
        \vdots & \vdots & \vdots
    \end{bmatrix}}_A \begin{bmatrix}
        a\\
        b\\
        c
    \end{bmatrix}=
    \underbrace{\begin{bmatrix}
        \Check{h}_{x_{i_1}, y_{j_1}} \\
        \Check{h}_{x_{i_2}, y_{j_2}}\\
        \vdots 
    \end{bmatrix}.}_B\\
\end{equation}
The parameters of the best-fit plane are found using the least squares method, given by
\begin{equation}
    \begin{bmatrix}
        a_{ls}\\
        b_{ls}\\
        c_{ls}
    \end{bmatrix}
    =(A^TA)^{-1}A^TB.
\end{equation}
Given Equation \eqref{eq:plane}, the slope at cell $s_{x_m,y_n}$ is given by 
\begin{equation}
    \label{eq:maxSlope}
    s_{x_m,y_n}=\sqrt{a_{ls}^2+b_{ls}^2}\text{,}
\end{equation} 
where $a_{ls}$ is the slope along the $x$ axis and $b_{ls}$ is the slope along the $y$ axis. 
 
\subsection{Generation of 2D Occupancy Maps for UAVs and UGVs}
\label{sec:map_generation}
In this subsection, two occupancy maps $\mathcal{M}=\{m_{x_i,y_j}\}_{M \times N}$ 
are constructed: one for UAVs ($\mathcal{M}_A$) and the other for UGVs ($\mathcal{M}_G$). 

\subsubsection{UAV map $\mathcal{M}_A$} The map generation begins by representing all cells of the 2D occupancy map with free space identified in subsection \ref{sec:FreeSpace} as free, i.e., $m_{x_i,y_j}=0$, when $F_{x_i,y_j}\neq \{\varnothing\}$ i.e, when the set $F_{x_i,y_j}$ is non-empty. The remaining cells $m_{x_i,y_j}$ that are not free, but have at least one neighboring cell that is free are examined using the following criteria to decide their occupancy probabilities:
\begin{equation}
    m_{x_i,y_j}= \left\{\begin{matrix}
        O_{cc}(x_i,y_j) \; & \text{if} \; O_{cc}(x_i,y_j) \geq O_{min} \\
        -1 \; &\text{otherwise}
    \end{matrix} \right.
\end{equation}
where $O_{cc}(x_i,y_j)$ is the occupancy value in the range $0$ and $1$ for the examined map cell $m_{x_i,y_j}$ computed using Equation \eqref{eq:occ} and $O_{min}$ is the minimum allowed occupancy value. If $O_{cc}(x_i,y_j) < O_{min}$, the cell's occupancy is deemed unknown and the value -1 is assigned. The function $O_{cc}(x_i,y_j)$, which calculates the overlapping percentage between the occupied ranges of $O_{x_i,y_j}$ and the neighboring free space in the height map $\mathcal{H}$, is defined as 
\begin{equation}
\label{eq:occ}
    O_{cc}(x_i,y_j)=\max_{{x_m,y_n}\in N(x_i,y_j)} \left( \frac{\sum_{o_{i} \in O_{x_i,y_j}} k(h_{x_m,y_n},o_{i})}{||h_{x_m,y_n}||} \right),
\end{equation}
where the function $k(h_{x_m,y_n},o_i)$ defined as: 
\begin{equation}
    \label{eq:overlap}
    \begin{matrix}
        k(h_{x_m,y_n},o_{i})=
        \max(0,\min(\hat{h}_{x_m,y_n},\hat{o}_{i})-\max(\Check{h}_{x_m,y_n},\Check{o}_{i}))
    \end{matrix}
\end{equation}
returns the length of the overlap between the two ranges, and when there is no overlap, the function returns $0$. If the examined cell has multiple neighboring free cells, only the free cell that results in the maximum occupancy is used. In Equation \eqref{eq:occ}, $N(x_i,y_j)$ is a set of neighboring cells $\{x_r, y_s\}$ with $m_{x_r,y_s}=0$, i.e., the cells containing free navigable range that satisfy the conditions in the following equation:
\begin{equation}
    \begin{aligned}
    N(x_i,y_j)= & \;\;\;\\
    \{ \{x_r,y_s\} | \max &(|x_r-x_i|,|y_s-y_j|) = 1, m_{x_r,y_s}=0 \}.    
    \end{aligned}
\end{equation}
Note that low walls and scattered objects are not relevant for aerial robot navigation, as there is navigable space for aerial robots above the walls and scattered objects. The approach proposed for the generation of $\mathcal{M}_A$ is concerned with the detection of free space for aerial navigation and ignores such entities for the aerial robot maps. 

\subsubsection{UGV map $\mathcal{M}_G$} The 2D map generation for ground-based robots utilizes the same free space and occupancy detection as the map generated for the UAV, but in addition, the slope map generated in subsection \ref{sec:slope} is used as well. Cells near low walls and scattered objects exhibit increased slope values as the height increases suddenly. 
In the UGV map $\mathcal{M}_G$, if the slope associated with a cell that is free is greater than $R_{MS}$, the cell is instead considered occupied, where $R_{MS}$ is the maximum slope that a ground robot can traverse. Thus, the low walls and scattered objects that are relevant for UGVs are detected and represented in the UGV map $\mathcal{M}_G$.
%
%
\section{2D to 3D Path Conversion} \label{sec:path}
As discussed in Section \ref{sec:introduction}, one of the primary advantages of the proposed method is its ability to plan paths using a 2D map instead of relying on a full 3D voxel map. Let us assume that a classical or state-of-the-art 2D path planner is used to find paths in the UAV and UGV maps. To make the 2D paths practicable for robots, especially UAVs, it is essential to determine the height, the $Z$ position, of each point along the 2D path. To address this challenge, the following process is employed to determine the $Z$ position of each point on a 2D path $\mathcal{P}$ generated on the 2D occupancy maps.

To determine the height $\mathcal{P}_{iZ}$ of each point $\mathcal{P}_i$ on the path $\mathcal{P}$, the following equation is used:
\begin{equation}
    \mathcal{P}_{iZ} = \max \left( \Check{\mathcal{H}}(\mathcal{P}_{i-p_f}),\dots, \Check{\mathcal{H}}(\mathcal{P}_{i+p_f})\right) + R_{off}
\end{equation}
where $\Check{\mathcal{H}}(\mathcal{P}_{i})$ represents the floor height in the height map $\mathcal{H}$ at the position of $\mathcal{P}_{i}$. The parameter $p_f$ is user-defined, specifying the number of steps that are considered along the path in front of and behind the current position of the robot. The parameter $R_{off}$ is another user-defined parameter that sets a desired safe height above the ground for the path. The use of the $\max$ function ensures that the path proactively adjusts its height $p_f$ steps before encountering an obstacle.
For UAVs, considering just the points along the path for height computation may prove insufficient to ensure safe distances from the objects in the neighborhood of the path. Therefore, when generating 3D paths for the UAVs, an additional check is performed to avoid collisions with a sphere-shaped safety region around the UAVs. 
The safety region has a radius of $0 < R_r \leq R_{maxZ}/2$. If the height calculated in the previous step results in a collision with either the floor or the ceiling in the height map $\mathcal{H}$, the point in the path that results in a collision is moved to a $Z$ position that does not cause a collision, ignoring $R_{off}$ at that point in the path. 
%
%
\section{Use Cases and Utility of the Proposed Method}\label{sec:useCases}
The authors identify several use cases for the 3D to 2D map conversion method proposed in this work. To begin with, a 2D map on a computer screen can be more intuitive for a human compared to a 3D map, to both visualize and interact with, for example, in selecting goal points or monitoring the navigation of a robot. 
The second use case arises in using a 2D map for global navigation for UAVs. Path planning using the 2D map is more efficient with respect to memory and CPU usage, as the navigation problems are quicker to solve since the $Z$-axis need not be considered. 
The third utility of the proposed methods is in enabling the use of methods (such as map segmentation or several 2D navigation solutions) that are designed specifically for 2D maps or scale poorly for a 3D map, to be applicable in 3D scenarios.

Map regions, objects, and paths found using the 2D maps can then be moved to 3D space using the height information of the environment extracted by the proposed method. This can be done using the method presented in Section \ref{sec:path}, which converts 2D paths planned on a  2D map to 3D paths for both UGVs and UAVs. 

The final utility arises in scenarios where multiple robotic agents need to communicate and exchange large maps with each other. Since the 2D map has a smaller file size than the full 3D Voxel map (as shown in Table \ref{tab:fileSize} in Section \ref{sec:validation}), it can be communicated between robots or to a remote location more rapidly within the allocated bandwidth. Although the 2D map and height map do not perfectly represent the environment, they still offer a good practical sense of the layout, enabling the robots to perform general path planning.
\section{Validation}\label{sec:validation}
The proposed method was validated using two 3D voxel maps of real environments gathered from robotic experiments: a small section of a cave and a large outdoor environment. A voxel map using the UFOMaps \cite{Duberg2020} was generated with a resolution of $\mathcal{V}_{res}=0.1$ m in the cave environment and with a resolution of $\mathcal{V}_{res}=0.2$ in the outdoor environment. 
Using the map conversion method outlined in Section \ref{sec:method}, 2D occupancy maps $\mathcal{M}_A$ and $\mathcal{M}_G$ were generated from voxel maps of the two environments. 

All scenarios were executed, and computation times were measured on the same computer. The tests were run on a single thread of an AMD 5850U CPU using Linux kernel version 6.9.7.
The parameters used by the method are listed in Table \ref{tab:setings}.
\begin{table}[H]
\centering
\caption{Parameters used in the validation cases.}
\resizebox{\linewidth}{!}{%
\begin{tabular}{|c|c|c|c|c|c|c|c|}
\hline
& $R_{maxZ}$ & $O_{min}$ & $S_A$& $R_{MS}$ & $R_{r}$ & $R_{off}$ & $p_f$ \\ \hline
UAV & 1 m & 0.5 & NA & 2 & 0.5 m & 1 m & 2 m / $\mathcal{V}_{res}$\\ \hline
UGV & 1 m & 0.5 & 0.2 & 2 & NA & 0.1 m & 0.5 m / $\mathcal{V}_{res}$\\ \hline
\end{tabular}}
\label{tab:setings}
\end{table}
\subsection{Conversion of static 3D voxel maps}
\begin{figure*}[htbp]
    \centering
    \begin{subfigure}[b]{.47\linewidth}
         \centering
         \includegraphics[width=\linewidth]{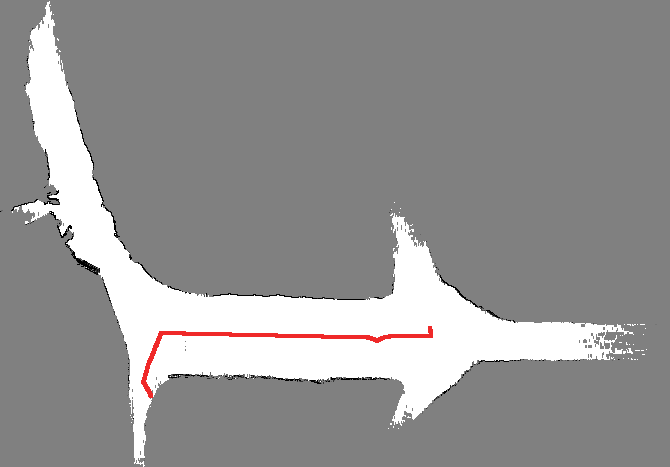}
         \caption{UAV map and path plan for a UAV}
         \label{fig:mapUAV}
     \end{subfigure}
     \hfill
     \begin{subfigure}[b]{.47\linewidth}
         \centering
         \includegraphics[width=\linewidth]{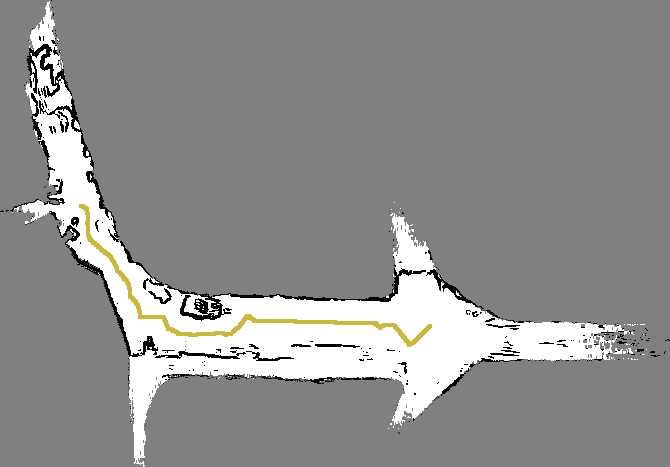}
         \caption{UGV map and path plan for a UGV}
         \label{fig:mapGR}
     \end{subfigure}
     \begin{subfigure}[b]{.47\linewidth}
         \centering
         \includegraphics[width=\linewidth]{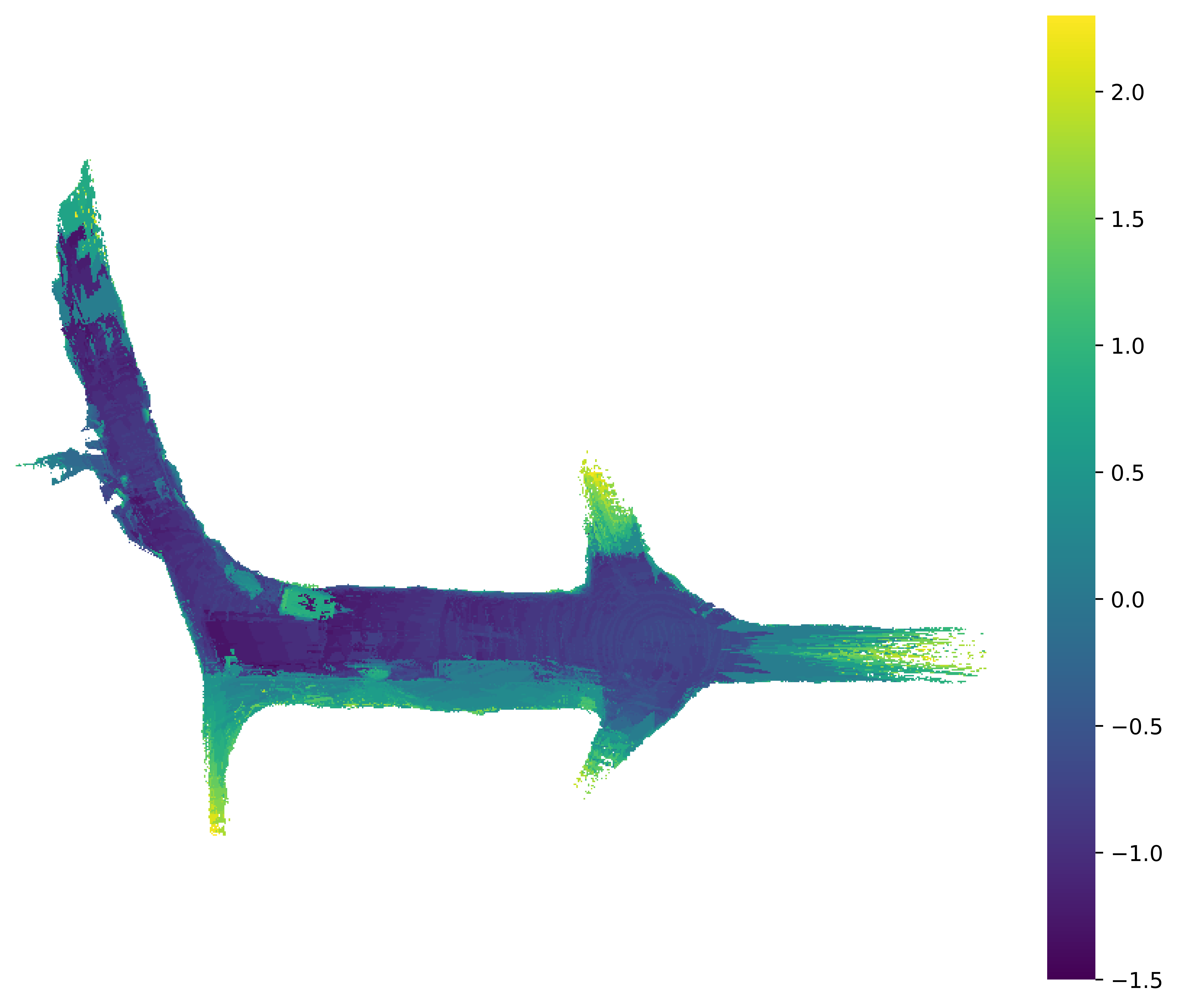}
         \caption{Height map}
         \label{fig:mapHight}
     \end{subfigure}
     \hfill
     \begin{subfigure}[b]{.47\linewidth}
         \centering
         \includegraphics[width=\linewidth]{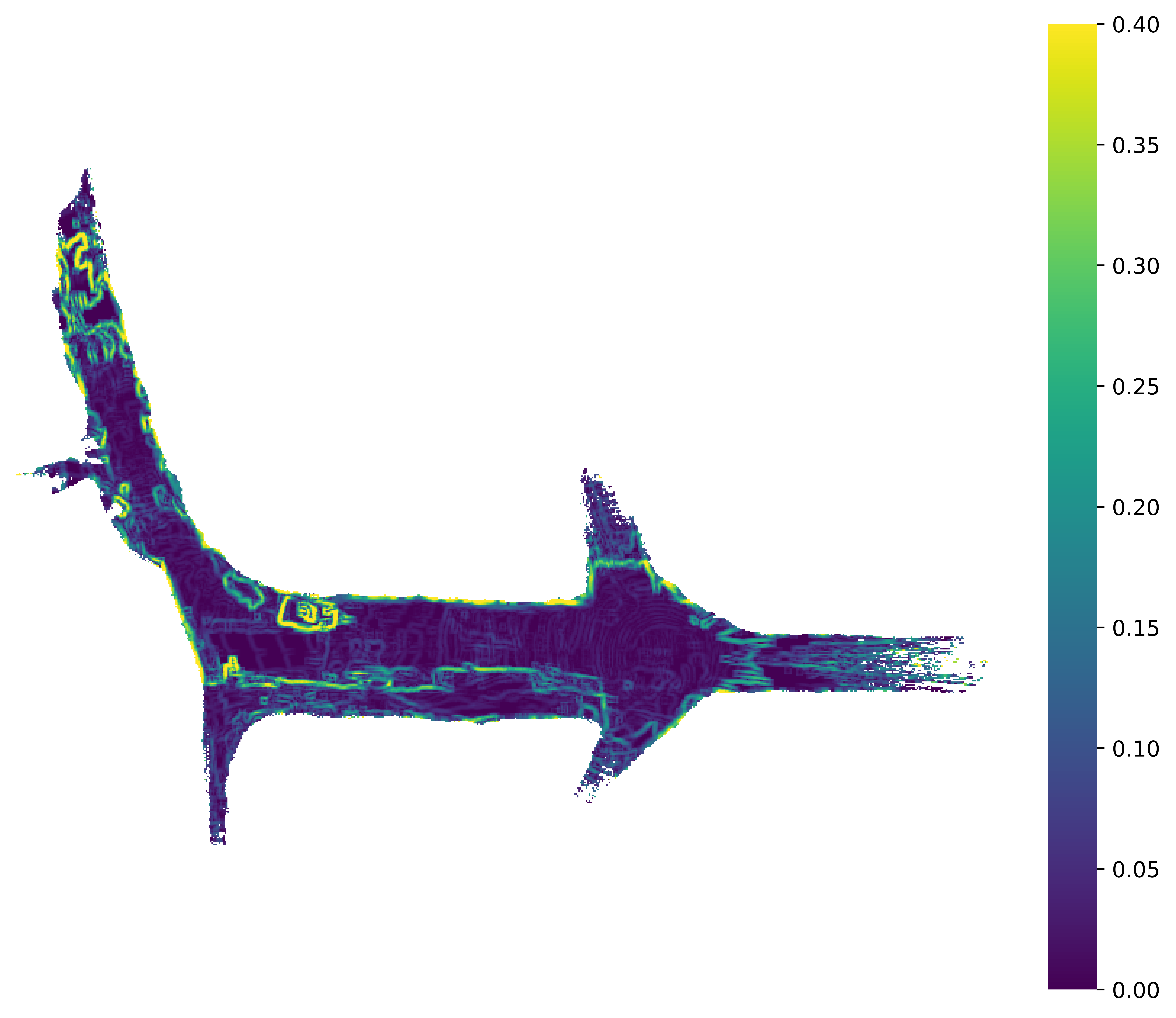}
         \caption{Slope map}
         \label{fig:mapSlope}
     \end{subfigure}
     
     \vspace{0.15cm}
     \begin{subfigure}[b]{.47\linewidth}
         \centering
         \includegraphics[width=\linewidth]{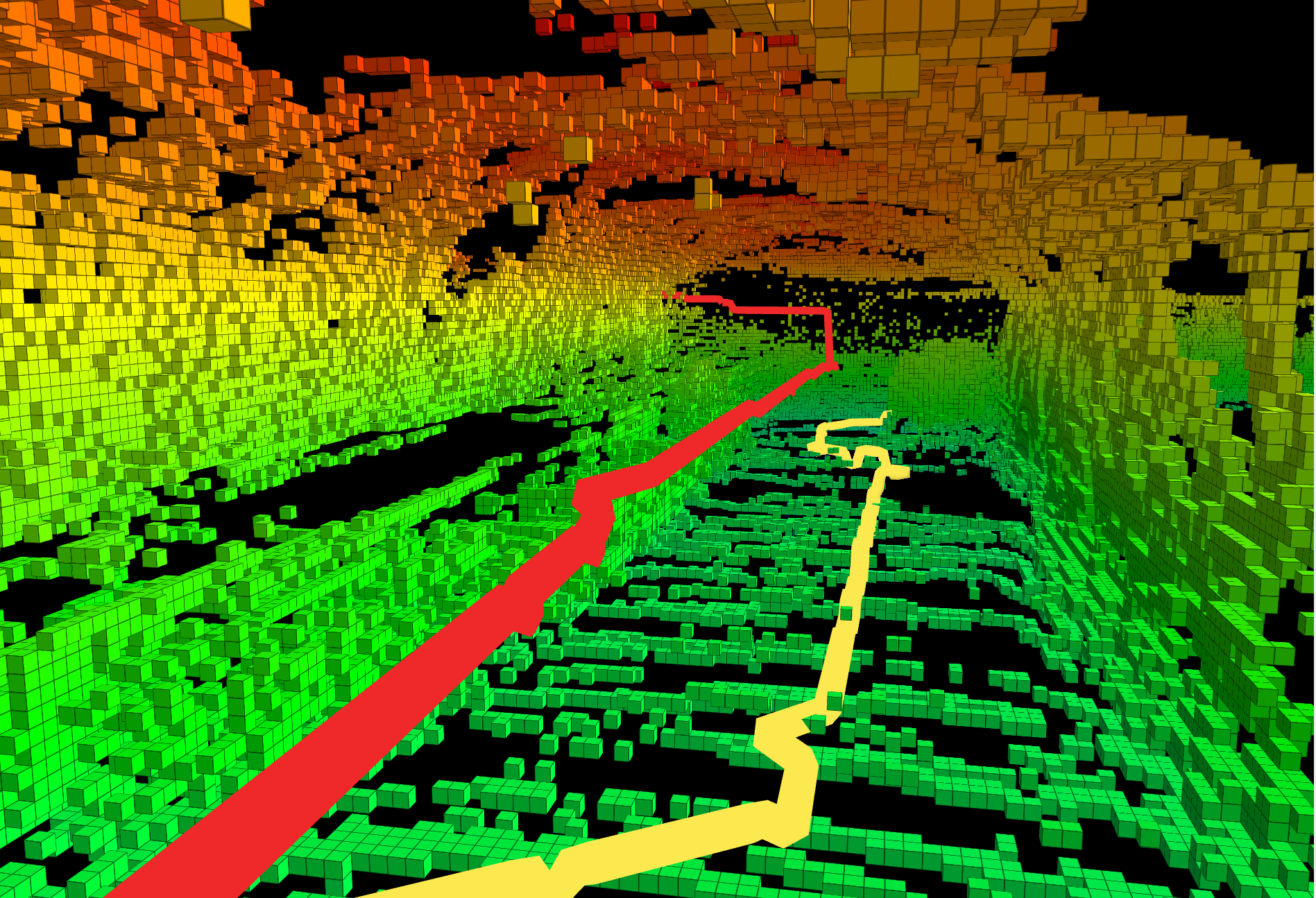}
         \caption{Voxel map perspective 1 with 3D paths for UAV (red) and UGV (yellow).}
         \label{fig:caveV1}
     \end{subfigure}
     \hfill
     \begin{subfigure}[b]{.47\linewidth}
         \centering
         \includegraphics[width=\linewidth]{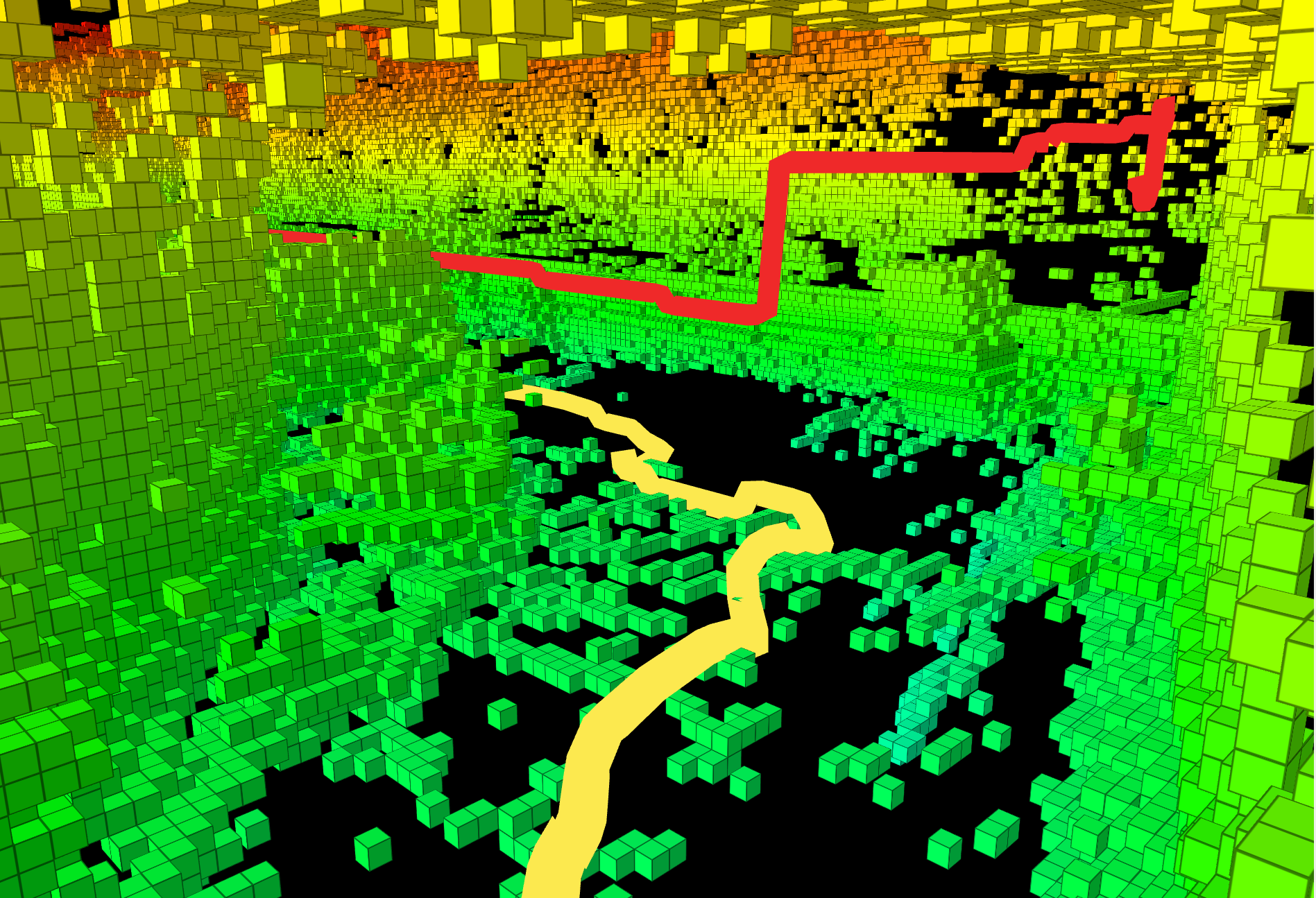}
         \caption{Voxel map perspective 2 with 3D paths for UAV (red) and UGV (yellow).}
         \label{fig:caveV2}
     \end{subfigure}
    \caption{2D maps and paths generated by the proposed methods for UAV and UGV in the cave environment. The results from the 2D path to 3D path conversion are shown in subfigures e) and f).}
    \label{fig:generatedMaps}
\end{figure*}

\begin{figure*}[htbp]
    \centering
    \begin{subfigure}[b]{.47\linewidth}
         \centering
         \includegraphics[width=\linewidth]{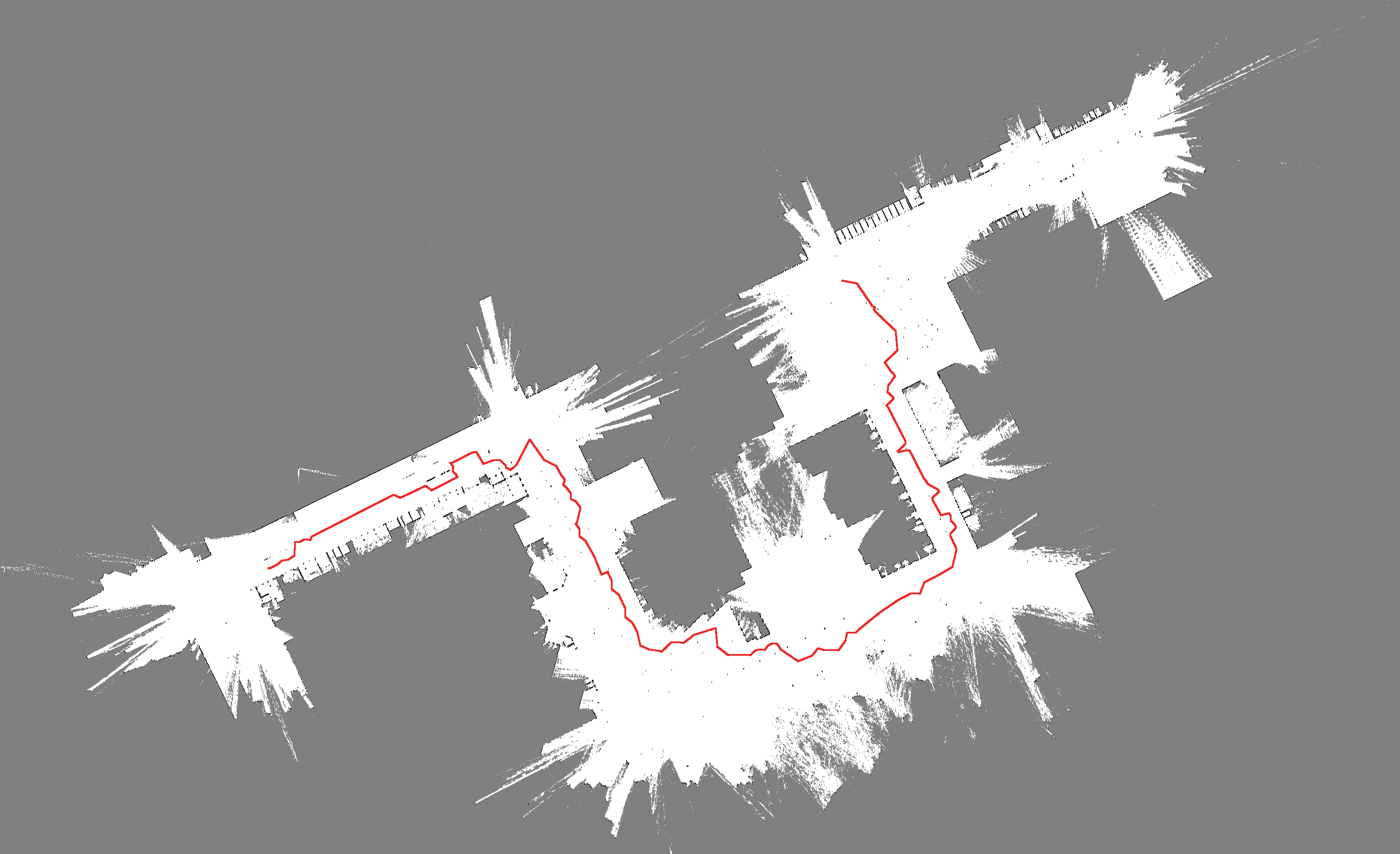}
         \caption{UAV map and path plan for a UAV}
         \label{fig:mapUAV_outdoor}
     \end{subfigure}
     \hfill
     \begin{subfigure}[b]{.47\linewidth}
         \centering
         \includegraphics[width=\linewidth]{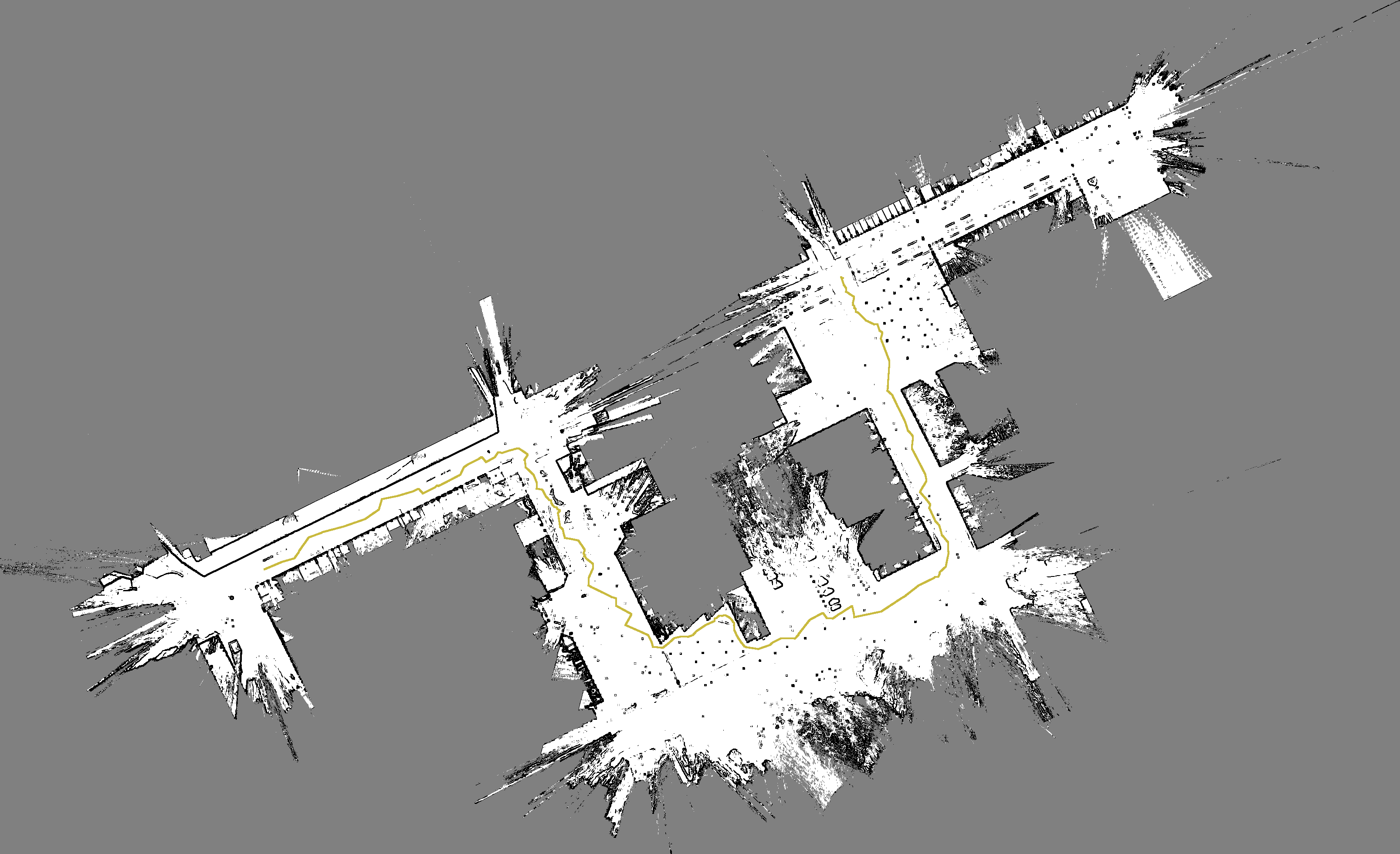}
         \caption{UGV map and path plan for a UGV}
         \label{fig:mapGR_outdoor}
     \end{subfigure}
     \begin{subfigure}[b]{.47\linewidth}
         \centering
         \includegraphics[width=\linewidth]{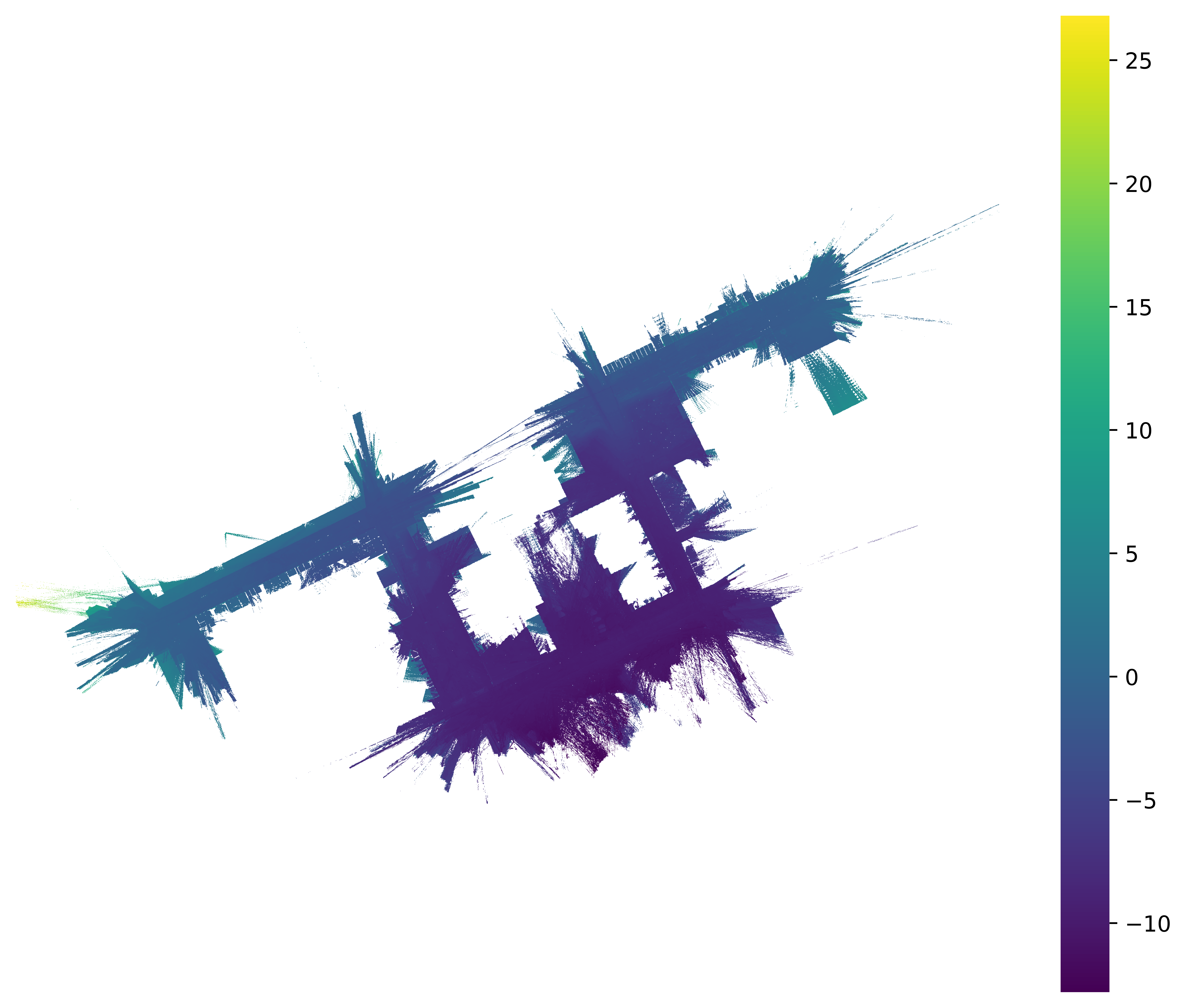}
         \caption{Height map}
         \label{fig:mapHight_outdoor}
     \end{subfigure}
     \hfill
     \begin{subfigure}[b]{.47\linewidth}
         \centering
         \includegraphics[width=\linewidth]{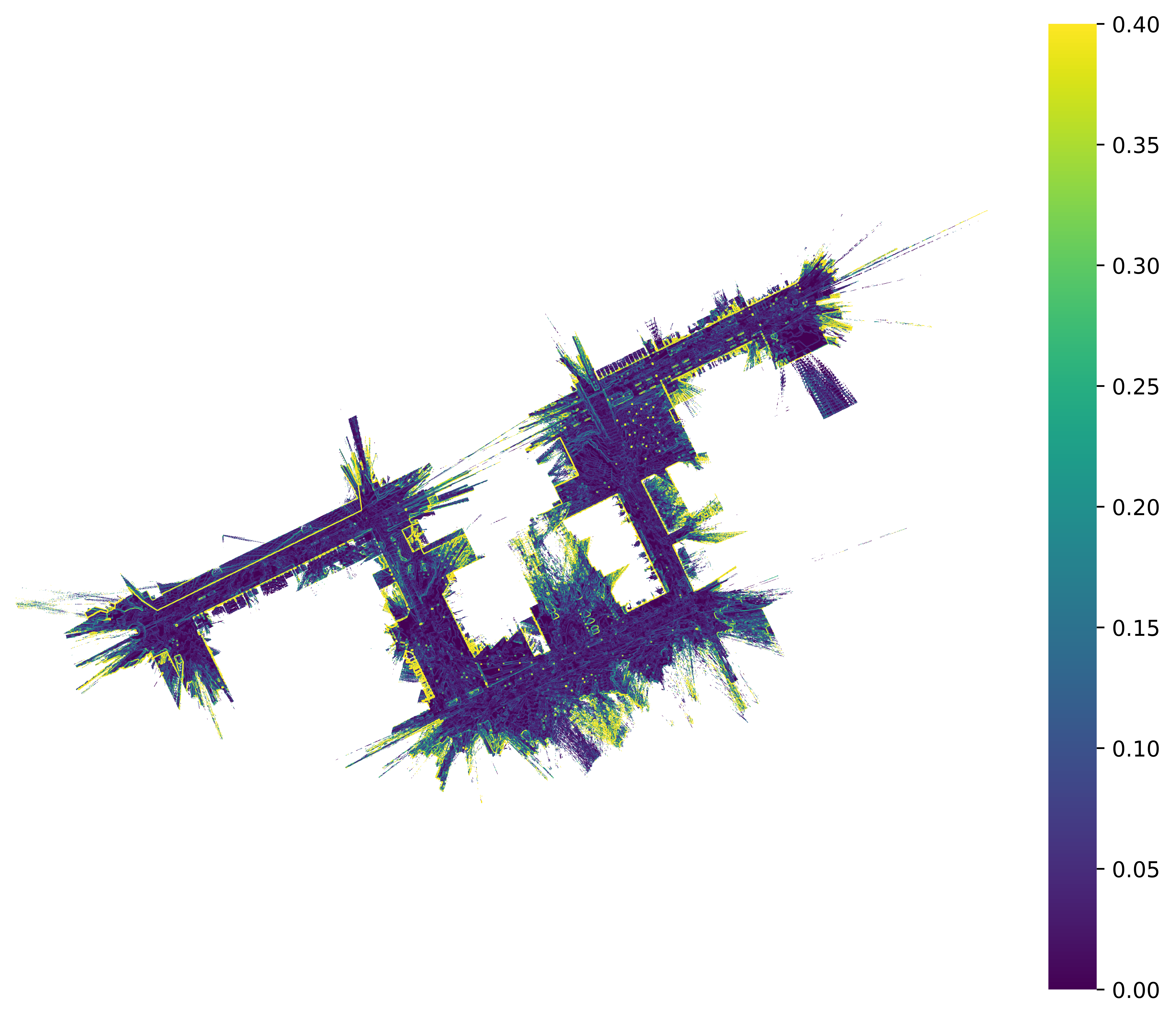}
         \caption{Slope map}
         \label{fig:mapSlope_outdoor}
     \end{subfigure}
     
     \vspace{0.15cm}
     \begin{subfigure}[b]{.47\linewidth}
         \centering
         \includegraphics[width=\linewidth]{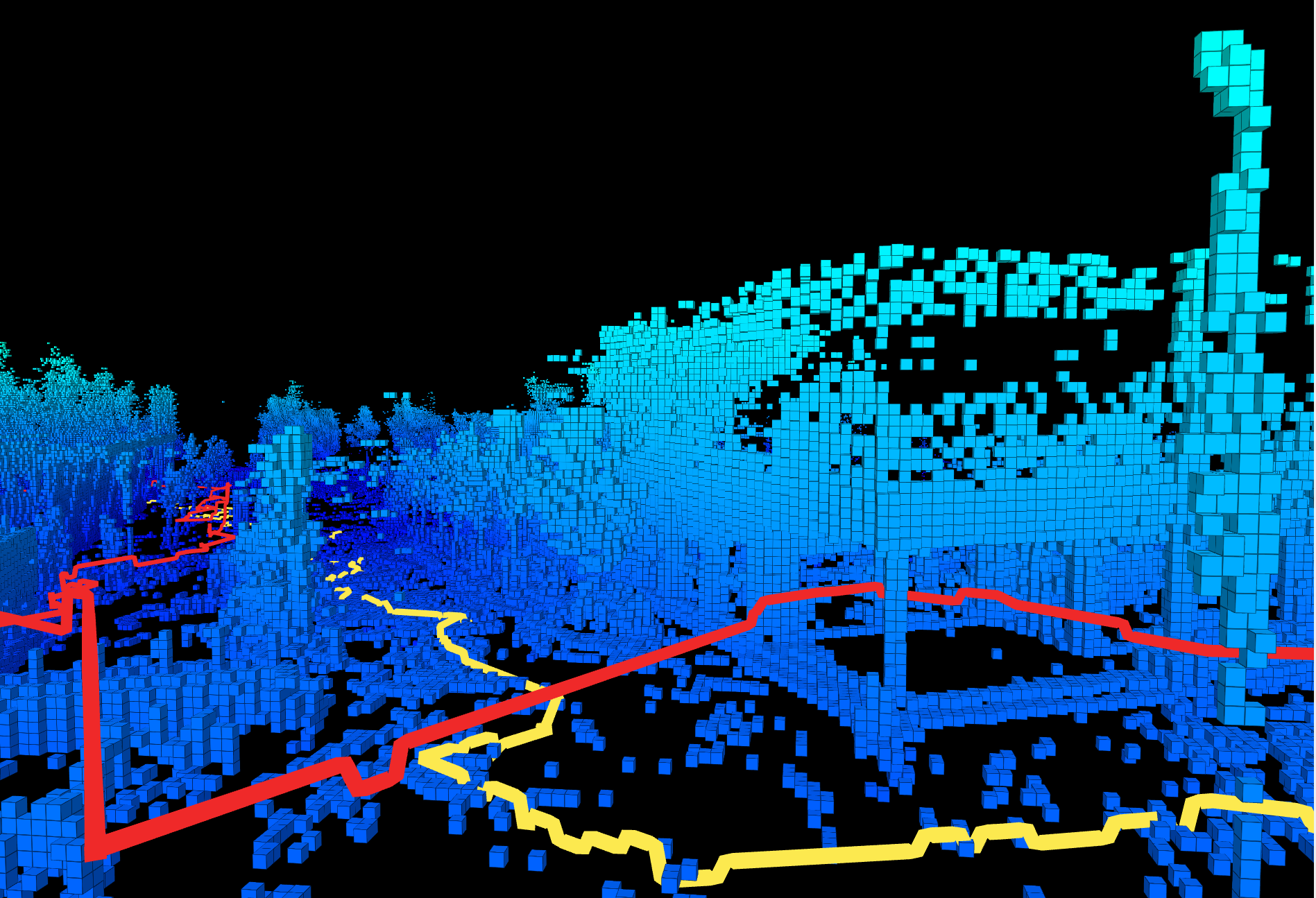}
         \caption{Voxel map perspective 1 with 3D paths for UAV (red) and UGV (yellow).}
         \label{fig:outdoorV1}
     \end{subfigure}
     \hfill
     \begin{subfigure}[b]{.47\linewidth}
         \centering
         \includegraphics[width=\linewidth]{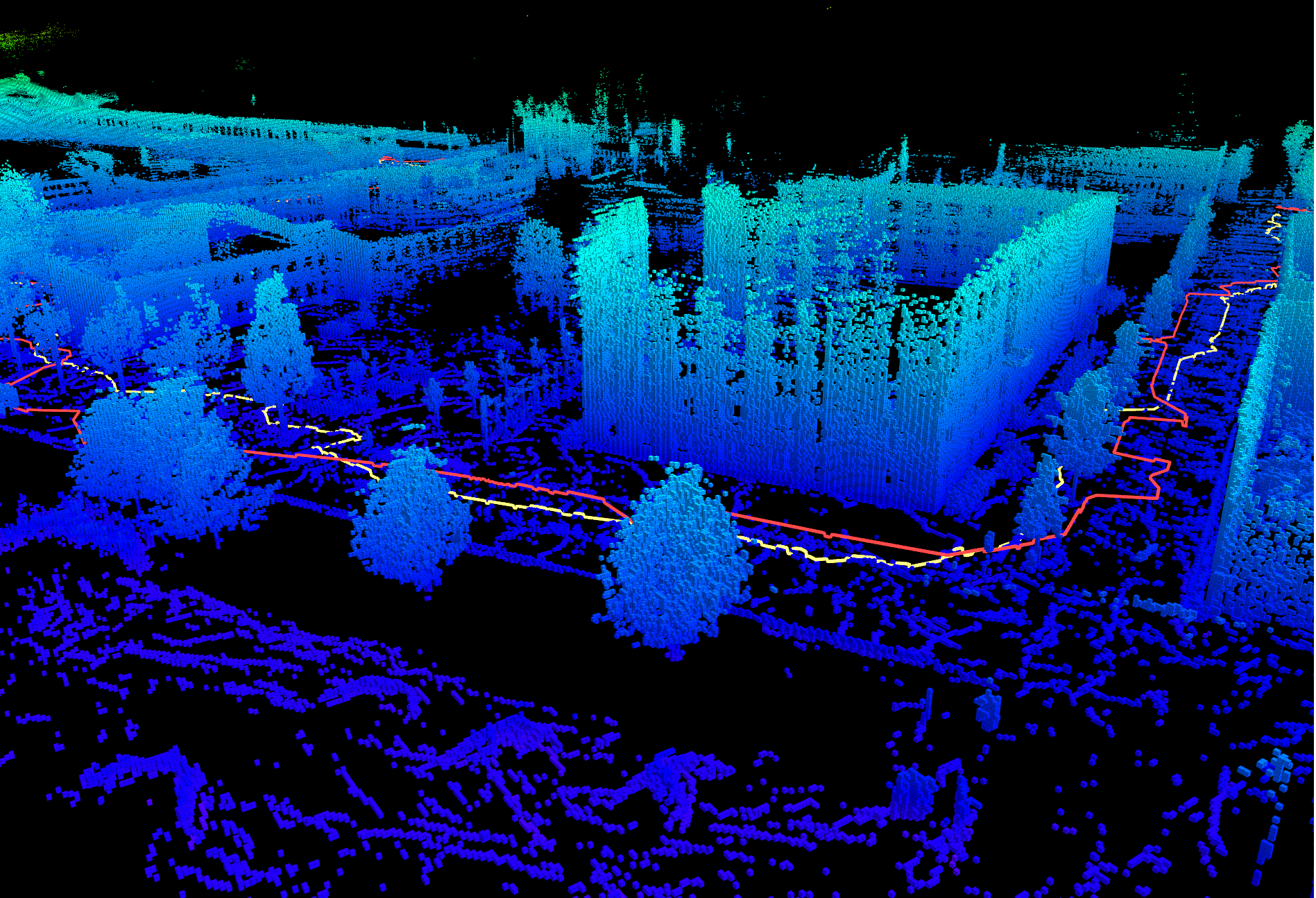}
         \caption{Voxel map perspective 2 with 3D paths for UAV (red) and UGV (yellow).}
         \label{fig:outdoorV2}
     \end{subfigure}
    \caption{2D maps and paths generated by the proposed methods for UAV and UGV in the cave environment. The results from the 2D path to 3D path conversion are shown in subfigures e) and f).}
    \label{fig:generatedMaps_outdoor}
\end{figure*}

The 2D maps for the Unmanned Aerial Vehicle (UAV) are shown in Figures \ref{fig:mapUAV} and \ref{fig:mapUAV_outdoor}, and for the Unmanned Ground Vehicle (UGV) are shown in Figures \ref{fig:mapGR} and \ref{fig:mapGR_outdoor}. 
Note that low walls and scattered objects are not relevant for aerial robots, as there is navigable space for aerial robots above the walls and scattered objects. 
For the ground robots, however, the walls and scattered objects are highly relevant and are retained by the proposed method, using the information contained in the slope map (the procedure is described in Section \ref{sec:map_generation}) and using the slope threshold parameter $R_{MS}$. The slope maps are depicted in Figures \ref{fig:mapSlope} and \ref{fig:mapSlope_outdoor}.

The slope map is generated from the height map $\mathcal{H}$. The height of the floor, $\Check{\mathcal{H}}$, with respect to the origin of the voxel map, is depicted in Figures \ref{fig:mapHight} and \ref{fig:mapHight_outdoor}.  

The computation time to perform the full map conversion was {\bf 0.23s} for the cave environment and {\bf 6.07s} for the outdoor environment.

Table \ref{tab:fileSize} shows the reduction in the raw size of one of the 2D maps with height data compared to the original voxel map. The voxel map is stored as an octree, while the 2D map and the height data are stored in a 2D matrix. 

To validate the path conversion from 2D to 3D using the 2D map $\mathcal{M}$ and the height map $\mathcal{H}$, two 2D paths were generated per environment: one for the Unmanned Aerial Vehicle (UAV), as shown in Figures \ref{fig:mapUAV} and \ref{fig:mapUAV_outdoor}, and another for the Unmanned Ground Vehicle (UGV), as seen in Figures \ref{fig:mapGR} and \ref{fig:mapGR_outdoor}.
The paths where generated using a path-planer developed for our earlier work on exploration~\cite{fredrikssonExploration} that is designed only for a 2D map. 
The 2D paths are converted into 3D paths for both the robots using the method in \ref{sec:path}. The resulting 3D paths are shown in the Figures \ref{fig:caveV1}, \ref{fig:caveV2}, \ref{fig:outdoorV1}, and \ref{fig:outdoorV2}. 
\begin{table*}[b]
\centering
\caption{Comparison between the raw sizes of different maps, expressed in megabytes (MB) and as a percentage compared to the original UFOMap used to generate the 2D map. The UFOMap is stored as an octree, while the 2D map and height data are stored in a 2D matrix.}
\begin{tabular}{|c|c|c|c|c|}
\hline
Scenario & UFOMap (MB) & 2D Map (MB) & 2D Map + Height Data (Floor) (MB) & 2D Map + Height data (Floor \& Ceiling) (MB) \\ 
\hline
Cave & 6.7 [100\%] & 0.3  [4.5\%] & 1.6 [23.9\%] & 2.9 [43.3\%] \\ \hline
Outdoor & 169.7 [100\%] & 6.3 [3.7\%] & 31.7 [18.7\%]& 57.1 [33.6\%] \\ \hline
\end{tabular}
\label{tab:fileSize}
\end{table*}
\subsection{Map Conversion During an Exploration Mission} \label{subsec:Exploration}
\begin{figure}
    \centering
    \begin{subfigure}[b]{.45\linewidth}
         \centering
         \includegraphics[width=\linewidth]{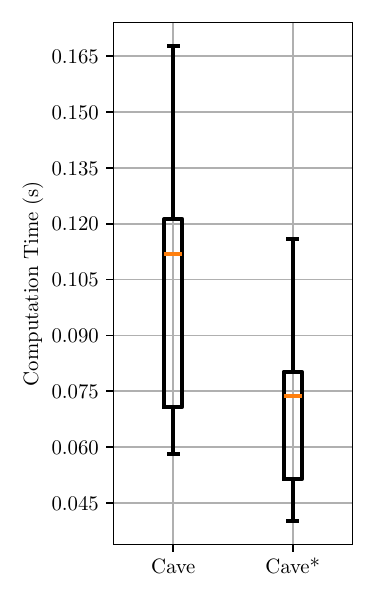}
         \caption{}
         \label{fig:timeCave}
     \end{subfigure}
     \begin{subfigure}[b]{.45\linewidth}
         \centering
         \includegraphics[width=\linewidth]{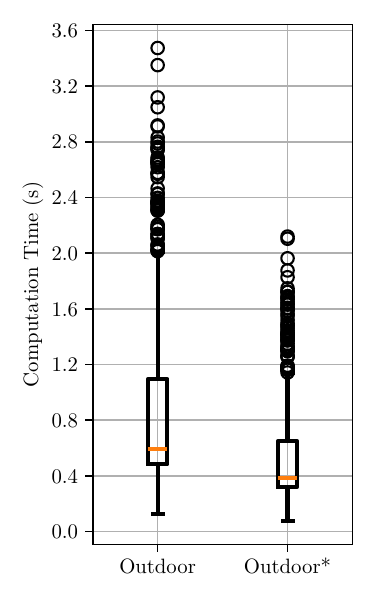}
         \caption{}
         \label{fig:timeOutdor}
     \end{subfigure}
    \caption{The computation time to update the 2D map generated from the voxel map under two exploration missions. The scenarios marked with '*' are the time required by the method, excluding the time needed to read the UFOmap.}
    \label{fig:timeC}
\end{figure}

\begin{figure}
    \centering
    \begin{subfigure}[b]{\linewidth}
         \centering
         \includegraphics[width=0.9\linewidth]{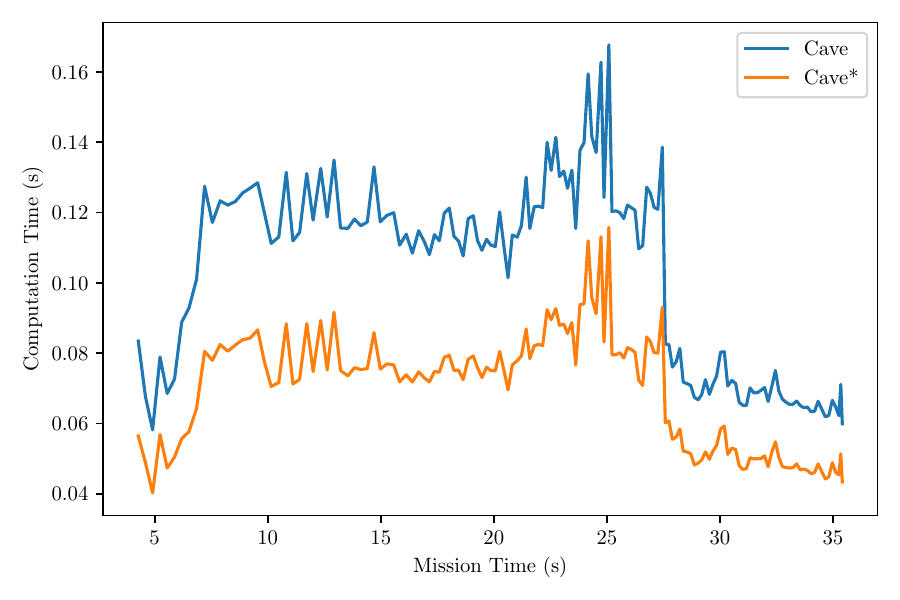}
         \caption{}
         \label{fig:timeCave2}
     \end{subfigure}
     \begin{subfigure}[b]{\linewidth}
         \centering
         \includegraphics[width=0.9\linewidth]{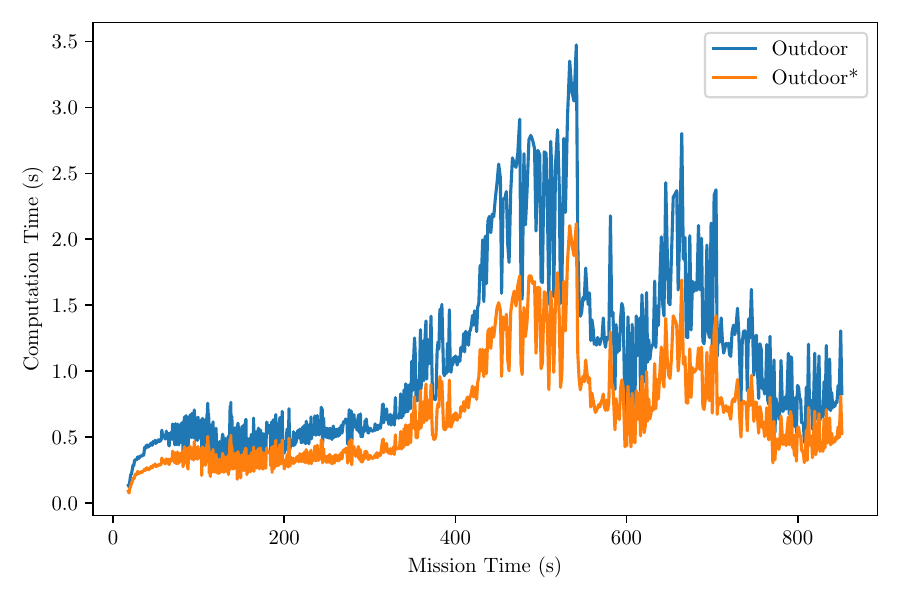}
         \caption{}
         \label{fig:timeOutdor2}
     \end{subfigure}
    \caption{The computation time to update the 2D map generated from the voxel map over the total mission time, under two exploration missions. The scenarios marked with '*' are the time required by the method, excluding the time needed to read the UFOmap.}
    \label{fig:timeC2}
\end{figure}

In each scenario, a robot platform explores the environment, and the 2D map is updated alongside the Voxel map. During the mission, only the section of the 3D map that was updated was used for map conversion and then to update the 2D map. 
The computation times for the two scenarios are presented in the box plots in Figure \ref{fig:timeC}. 
There are two times displayed in the box plot. Scenarios marked with '*' exclude the time spent reading the voxel map produced by UFOMap. Since this time will vary depending on the mapping framework used with the method, the time marked with '*' more accurately reflects the time used by the conversion method. In Figure \ref{fig:timeC2}, computation time relative to the total mission time of  the exploration is shown.
  
%
%
\section{Discussion} \label{sec:discussion}
As demonstrated in Section \ref{sec:validation}, the method effectively converts voxel maps in environments featuring non-uniform height variations and in the presence of overhangs/ceilings. The UGV map includes obstacles and walls impassable by the UGV, thus ensuring comprehensive environmental representation. In existing conversion methods \cite{Hornung2013}\cite{Yang2018} for voxel-based maps, this would not have been possible, especially in outdoor scenarios where the height difference is more than 10 meters between the starting position's ground level and the map's lowest point. 
Another advantage of the proposed map conversion is in the raw file size of the generated 2D maps compared to the original voxel map, as can be seen in Table \ref{tab:fileSize}. In cases where only UAV or UGV maps need to be transferred, the required size is less than $5\%$ of the original voxel map, which is significant in multi-agent scenarios where bandwidth is limited. 

The time required to convert a static map increases with map size. The conversion time for the small cave scenario is 0.23s, whereas a larger outdoor scenario requires 6.7s. This increase is not an issue since the conversion of the full map is generally performed only once. 
In dynamic scenarios where the map size changes or certain areas are updated, such as during an exploration mission (presented in section \ref{subsec:Exploration}), only the modified sections of the map need to be updated.
In smaller enclosed environments, like a subterranean environment, the average computation time is only 0.11s, which allows for near real-time map updates, as seen in Figure \ref{fig:timeCave}. In contrast, in more open environments, such as the outdoor scenario, the conversion time increases to an average of 0.5s.
Figure \ref{fig:timeCave2} illustrates that the computation time in the subterranean environment remains stable because the tunnels have approximately constant width. Conversely, in the outdoor scenario shown in Figure \ref{fig:timeOutdor2}, the robot explores more open areas during the second half of the exploration mission, resulting in significant spikes in computation time.

An issue with the UGV map generated by the method is the inaccurate representation at some frontiers, which are the boundaries between free and unknown space. 
This inaccuracy arises from the greedy method for floor detection presented in Section \ref{sec:FreeSpace}, which inaccurately classifies the floor height around the frontiers, resulting in a steep slope in these areas.
Despite this limitation, the mapped area remains accurate for UGV navigation. In scenarios such as autonomous exploration, reliable frontier detection is an absolute requirement, and this necessitates reliance on the voxel map for frontier detection.

A limitation of using a 2D planner for UAV path planning instead of a full 3D planner, in scenarios with multiple 3D paths leading to a target point, is that a 2D planned may choose sub-optimal paths, such as flying over a tall obstacle instead of circumventing it. A potential solution could be integrating height differences into the 2D planner's calculations, thus utilizing the height map produced by the method more effectively.
%
%
%
%
%
%
\section{Conclusions} \label{sec:conclusion}
Motivated by the utility of converting 3D maps to 2D for global navigation, which enables the use of methods designed for 2D map in 3D environments, and faster information sharing between robots, this work presented a novel method for converting 3D voxel maps to 2D occupancy maps for UAVs and UGVs, along with accompanying height and slope maps. 
A method was also proposed for converting the paths generated by the 2D planer back into 3D space using the height information extracted by the method. The method proposed in this paper is capable of generating 3D paths for both ground and aerial robots and was validated successfully in two different environments (a cave and an outdoor scenario).


\bibliographystyle{./IEEEtranBST/IEEEtran}
\bibliography{./IEEEtranBST/IEEEabrv,references}

\begin{thebibliography}{10}
\providecommand{\url}[1]{#1}
\csname url@rmstyle\endcsname
\providecommand{\newblock}{\relax}
\providecommand{\bibinfo}[2]{#2}
\providecommand\BIBentrySTDinterwordspacing{\spaceskip=0pt\relax}
\providecommand\BIBentryALTinterwordstretchfactor{4}
\providecommand\BIBentryALTinterwordspacing{\spaceskip=\fontdimen2\font plus
\BIBentryALTinterwordstretchfactor\fontdimen3\font minus \fontdimen4\font\relax}
\providecommand\BIBforeignlanguage[2]{{%
\expandafter\ifx\csname l@#1\endcsname\relax
\typeout{** WARNING: IEEEtran.bst: No hyphenation pattern has been}%
\typeout{** loaded for the language `#1'. Using the pattern for}%
\typeout{** the default language instead.}%
\else
\language=\csname l@#1\endcsname
\fi
#2}}

\bibitem{Megalingam2023}
R.~K. Megalingam, S.~Tantravahi, H.~S. S.~K. Tammana, and H.~S.~R. Puram, ``\BIBforeignlanguage{en}{{2D}-{3D} hybrid mapping for path planning in autonomous robots},'' \emph{\BIBforeignlanguage{en}{International Journal of Intelligent Robotics and Applications}}, vol.~7, no.~2, pp. 291--303, June 2023.

\bibitem{Wulf2004}
O.~Wulf, K.~Arras, H.~Christensen, and B.~Wagner, ``{2D} mapping of cluttered indoor environments by means of {3D} perception,'' in \emph{{IEEE} {International} {Conference} on {Robotics} and {Automation}, 2004. {Proceedings}. {ICRA} '04. 2004}, vol.~4, Apr. 2004, pp. 4204--4209 Vol.4, iSSN: 1050-4729.

\bibitem{Sandfuchs2021}
S.~Sandfuchs, M.~P. Heimbach, J.~Weber, and M.~Schmidt, ``Conversion of depth images into planar laserscans considering obstacle height for collision free {2D} robot navigation,'' in \emph{2021 {European} {Conference} on {Mobile} {Robots} ({ECMR})}, Aug. 2021, pp. 1--6.

\bibitem{Mora2023}
A.~Mora, R.~Barber, and L.~Moreno, ``Leveraging {3D} {Data} for {Whole} {Object} {Shape} and {Reflection} {Aware} {2D} {Map} {Building},'' \emph{IEEE Sensors Journal}, pp. 1--1, 2023.

\bibitem{Nam2017}
T.~H. Nam, J.~H. Shim, and Y.~I. Cho, ``\BIBforeignlanguage{en}{A 2.{5D} {Map}-{Based} {Mobile} {Robot} {Localization} via {Cooperation} of {Aerial} and {Ground} {Robots}},'' \emph{\BIBforeignlanguage{en}{Sensors}}, vol.~17, no.~12, p. 2730, Dec. 2017.

\bibitem{Hornung2013}
A.~Hornung, K.~M. Wurm, M.~Bennewitz, C.~Stachniss, and W.~Burgard, ``\BIBforeignlanguage{en}{{OctoMap}: an efficient probabilistic {3D} mapping framework based on octrees},'' \emph{\BIBforeignlanguage{en}{Autonomous Robots}}, vol.~34, no.~3, pp. 189--206, Apr. 2013.

\bibitem{Yang2018}
S.~Yang, S.~Yang, and X.~Yi, ``An {Efficient} {Spatial} {Representation} for {Path} {Planning} of {Ground} {Robots} in {3D} {Environments},'' \emph{IEEE Access}, vol.~6, pp. 41\,539--41\,550, 2018.

\bibitem{Li2024}
Y.~Li, D.~Wang, Q.~Li, G.~Cheng, Z.~Li, and P.~Li, ``\BIBforeignlanguage{en}{Advanced {3D} {Navigation} {System} for {AGV} in {Complex} {Smart} {Factory} {Environments}},'' \emph{\BIBforeignlanguage{en}{Electronics}}, vol.~13, no.~1, p. 130, Jan. 2024.

\bibitem{Kamarudin2013}
K.~Kamarudin, S.~M. Mamduh, A.~Y.~M. Shakaff, S.~M. Saad, A.~Zakaria, A.~H. Abdullah, and L.~M. Kamarudin, ``Method to convert {Kinect}'s {3D} depth data to a {2D} map for indoor {SLAM},'' in \emph{2013 {IEEE} 9th {International} {Colloquium} on {Signal} {Processing} and its {Applications}}, Mar. 2013, pp. 247--251.

\bibitem{Garrote2017}
L.~Garrote, J.~Rosa, J.~Paulo, C.~Premebida, P.~Peixoto, and U.~J. Nunes, ``{3D} point cloud downsampling for {2D} indoor scene modelling in mobile robotics,'' in \emph{2017 {IEEE} {International} {Conference} on {Autonomous} {Robot} {Systems} and {Competitions} ({ICARSC})}, Apr. 2017, pp. 228--233.

\bibitem{Brahmanage2019}
G.~Brahmanage and H.~Leung, ``Building {2D} {Maps} with {Integrated} {3D} and {Visual} {Information} using {Kinect} {Sensor},'' in \emph{2019 {IEEE} {International} {Conference} on {Industrial} {Cyber} {Physical} {Systems} ({ICPS})}, May 2019, pp. 218--223.

\bibitem{Yusefi2020}
A.~Yusefi, A.~Durdu, and C.~Sungur, ``\BIBforeignlanguage{en}{{ORB}-{SLAM}-based {2D} {Reconstruction} of {Environment} for {Indoor} {Autonomous} {Navigation} of {UAVs}},'' \emph{\BIBforeignlanguage{en}{Avrupa Bilim ve Teknoloji Dergisi}}, pp. 466--472, Oct. 2020.

\bibitem{MurArtal2015}
R.~Mur-Artal, J.~M.~M. Montiel, and J.~D. Tardós, ``{ORB}-{SLAM}: {A} {Versatile} and {Accurate} {Monocular} {SLAM} {System},'' \emph{IEEE Transactions on Robotics}, vol.~31, no.~5, pp. 1147--1163, Oct. 2015.

\bibitem{Gim2021}
H.~Gim, M.~Jeong, and S.~Han, ``Autonomous {Navigation} {System} with {Obstacle} {Avoidance} using 2.{5D} {Map} {Generated} by {Point} {Cloud},'' in \emph{2021 21st {International} {Conference} on {Control}, {Automation} and {Systems} ({ICCAS})}, Oct. 2021, pp. 749--752, iSSN: 2642-3901.

\bibitem{Duberg2020}
D.~Duberg and P.~Jensfelt, ``{UFOMap}: {An} {Efficient} {Probabilistic} {3D} {Mapping} {Framework} {That} {Embraces} the {Unknown},'' \emph{IEEE Robotics and Automation Letters}, vol.~5, no.~4, pp. 6411--6418, Oct. 2020.

\bibitem{fredrikssonExploration}
S.~Fredriksson, A.~Saradagi, and G.~Nikolakopoulos, ``Robotic exploration through semantic topometric mapping,'' in \emph{2024 IEEE International Conference on Robotics and Automation (ICRA)}, 2024, pp. 9404--9410.

\end{thebibliography}
\end{document}